# Communication of Social Agents and the Digital City – A Semiotic Perspective

Victor V. Kryssanov[1], Masayuki Okabe[1], Koh Kakusho[2], and Michihiko Minoh[2]

[1] Japan Science and Technology Corporation, Japan
[2] Center for Information and Multimedia Studies, Kyoto University,
Kyoto 606-8501, Japan
{kryssanov, okabe, kakusho, minoh}@mm.media.kyoto-u.ac.jp

**Abstract.** This paper investigates the concept of digital city. First, a functional analysis of a digital city is made in the light of the modern study of urbanism; similarities between the virtual and urban constructions are pointed out. Next, a semiotic perspective on the subject matter is elaborated, and a terminological basis is introduced to treat a digital city as a self-organizing meaning-producing system intended to support social or spatial navigation. An explicit definition of a digital city is formulated. Finally, the proposed approach is discussed, conclusions are given, and future work is outlined.

## 1 Introduction

A digital city may be very generally defined as a collection of digital products and information resources made of a large distributed database of heterogeneous documents of various digital genres – (hyper)texts, photographs, maps, animated images, and the like – deployed to provide services aimed at facilitating social and/or spatial navigation in a virtual (e.g. "information" or "communication") or physical (e.g. geographical) space. Paramount to any digital city mission is the ability to deliver information of interest in a timely manner to its users. To do this, digital cities exploit a computer network and a client-server protocol, allowing the user to browse across digital documents through appropriately ordered hyperlinks and retrieve information as needed. An effective digital city supports access to all its repositories of relevant knowledge and data in both the raw and quality-filtered forms, and there can be several search engines installed to carry out the retrieval process. Naturally, networking and information retrieval are considered key issues in the development of digital cities.

As part of the information delivery, a digital city seeks to enable uncomplicated and correct interpretation of the results of a user's query. Examples of this include but are not limited to: providing the user with the related context as an aid for understanding the results (or even the query itself, in the case of exploratory search), illustrating the results with a suitable metaphor or analogy, and utilizing feedback from the user (or some data about the user) to adjust the strategy for retrieving or displaying the information to make it more accessible and meaningful to the user (the so-called

"adaptive navigation support"). Another important issue that is thus immediately falling under the purview of digital city developers is human-computer interface.

Reflecting the present understanding of the concept of digital city, which is far from unified and is subject to discussion (and yet confusion), the literature abounds in technical descriptions of implemented and projected digital cities [18][1]. The authors typically approach the task of the development with a narrow focus, defining a digital city through its functions or even contents with vague terms, such as "useful information," "communication," "social agent," "community network," and the like, and with *ad hoc* "common sense" design decisions, which may have unpredictable (especially, on a long-term scale) consequences, and which are often of arbitrary relevance to the users' needs. Evidence of the latter can be found already in the very attempt to characterize digital products intended for different purposes (e.g. social *vs.* geographical navigation) and different types of users by utilizing the loose metaphor of city without clarifying which, if any, aspects or features of the city original concept – the material grounding, functionality, dynamics, structure, or other – are to be adopted. Besides, it is admitted that the inter-disciplinary theoretical study of digital city remains in its infancy and is currently of little help to the practical developers.

The presented work aims to establish a basis for scientific investigation of digital cities and explore fundamental properties of a digital city as an information structure. In this paper, a definition of a digital city as an organization of interacting social agents is introduced, based on a semiotic interpretation of a system-theoretic model of communication. The definition is to explicate the concept of digital city and to elaborate perspectives on the research and development for the future.

The rest of the paper is in four parts. The next section analyzes the concept of digital city. Section 3 develops a semiotic view on the subject matter. This is followed by a section that strives to define a digital city in a manner sufficiently precise for both the academic and practical needs. The final section then discusses the study theoretical findings, formulates some conclusions, and gives information on forthcoming research.

## 2 Concept of Digital City

### 2.1 Metaphor of City

It is evident that "digital city" is a metaphor. Metaphors (from Greek *metaphora* – transfer) serve to create new meanings by transferring the semantics of one concept into the semantics of another concept. Metaphors are habitually used to interpret an unknown "world" (perception, experience, etc.) – the target – in terms of a familiar world – the source. Metaphorical explanation often helps us understand highly abstract and complex phenomena by relating them to phenomena we know well (or, at least, better). In so doing, a metaphor preserves (part of) structure of the original concept, but substitutes its functional contents, anticipating the corresponding change in its properties and meaning.

---

[1] Throughout the paper, we will use this publication as a representative collection of studies of digital cities.

A metaphor can be expressed with worlds, gesture, in a graphical manner, or through behavior – essentially, metaphors are (combinations of) signs. To successfully apply a metaphor, one should understand not only the systemic organization of the source, but the rôle of the source's larger context that has to be realized and presented in the context intended for the target. (Proper) metaphors are not merely somehow convenient selections of signs. Rather, they are selections of consistent logical systems or theories, with which one can generate new meaningful signs from signs already existed [37]. It seems rational to assume that the logic of a digital city could root in the logic of real cities, if this metaphor is to be properly used. Another (perhaps, stronger) assumption can be made that digital cities as "virtual structures" have much of the properties inherent in the social communities.

Juval Portugali, in his survey of the study of urbanism, described a present-day city as a conglomerate of people together with their artifacts – buildings, roads, communications, etc. – that is "actually not a city but a text written by millions of unknown writers, unaware that they are writers, read by millions of readers, each reading his or her own personal and subjective story in this ever-changing chaotic text, thus changing and recreating and further complicating it" [30, 29]. A city is *complex*. It consists of numerous components, which interact, and which are created by (and from) other components, thereby continuously re-producing the fabric of the city. A city is *self-organizing*. It, like all self-organizing systems, exchanges the resources – matter, energy, and people – and information with its environment and is, in this sense, *open*. At the same time, a city is *closed* (to a degree) in the sense that its structure is determined internally (again, to a degree), and the environment does not control how the city organizes itself. In complex system theory, complete closure means that every component of a system is produced solely by other components of the same system without influence from the outside – a requirement hardly reachable in social or even biological organizations. However, once a city starts to distinguish itself from the environment by creating a *boundary*, it can achieve a sufficient degree of organizational closure to be seen (though, controversially [25]) as an *autopoietic system* that is a form of self-organization [24].

Due to its composite makeup and the complexity of internal interactions, a city is generally indescribable in terms of cause and effect or in terms of probabilities [7]. This makes it extremely difficult to study the city basic properties. Recently, however, some conceptual and mathematical approaches (such as dissipative structures, synergetics, and cellular automata), borrowed from the natural sciences, have successfully been applied to the study of cities with the focus on not to control or predict behaviors of city components, but to deliberately participate in and sensefully "shape" the city development by acting at the global, social and organizational level [30].

Turning now to the case of digital cities, one can quickly prove the openness and the complexity of the virtual organizations. Intuitively, digital cities should be open, since they constantly exchange information with their environments, and they are indeed complex, each comprising a number of dynamic information resources. Not so straightforward is the question of organizational closure: it is unclear what a digital city produces and how it reproduces. A still more difficult question is, how can a digital city separate itself from the environment and what is the boundary? By answering these, we would clarify whether and to what extent the logic of cities and social systems in general is applicable to an information web-structure called "digital city."

### 2.2 Navigation with Digital City

Even at the level of parts, self-organization does not really mean freedom but a controlled collective behavior towards achieving a common (for the entire system) goal in an environment [16]. There is a controlling mechanism "hidden" and distributed over all the system parts that determines a strategy for the system, which is usually implemented as more or less inclusive constraints imposed on each part's behavior. While the principal goal of any ("living") self-organizing system must imply its long-term survival in a variable environment, i.e. the maintenance of the system invariant structure – its *identity*, the tactical (or transient) goals usually determine the system parts' behavior in every local situation and at every particular time. To "survive" for a digital city would mean to uphold the stability of its structure with the designated functionality supporting (social, geographical, spatial, etc.) *navigation* despite environmental disturbances.

Perhaps most generally, navigation in an environment[2] was defined in [34] as a four-stage iterative process that includes: 1) *perception* of the environment, 2) reconciliation of the perception and cognition (i.e. *understanding*), 3) deciding whether the current goal has been reached (i.e. *decision-making*), and 4) choosing and performing the next action (i.e. *adjustment* of the behavior). Among these stages, the last two have a noticeably subjective character and are solely on the navigator's side, whereas the other two depend on "objectively" available – sensed – information about the environment. It is perception that first represents "raw" sensory data and provides for further interpretation by putting the resultant representations into a context of the scene perceived (e.g. by simply combining the representations together). When information obtained through the senses is not enough for establishing or re-establishing meanings of the environment necessary for successful decision-making, the navigator may ask for help a guide – someone, who could presumably know more about the environment. A digital city may be thought of as such a guide: in the navigation process, it works to enhance and complement the navigator's sensing capabilities. In other words, a digital city is to "produce" information about the navigation space.

Perceptual Control Theory [31] proposes an explanation of the control mechanism for complex self-organizing systems. The theory tells us that a perceiving system normally seeks to bring the perceived situation to its goal or preferred state by utilizing (negative) feedback from the environment: if the situation deviates from the goal, the system acts and adapts, possibly changing its own state and the state of the envi-ronment, and the new situation is again sensed and estimated in respect to the goal. The loop repeats and keeps the system in a stable goal-directed state, environmental perturbations and compensating actions notwithstanding. Although a digital city can, in principle, sense its environment directly (e.g. through cameras and transducers, as in the "Helsinki Arena 2000" project [22]), there is no other way for it to determine the context and, hence, semantics necessary for making the sensed information *meaningful* for the navigation process, but (ultimately) by drawing on expertise of its users and utilizing feedback from them. The users together with their knowledge can and in fact should be considered as indispensable and *constitutive* parts of the digital city.

---

[2] It should be noted that for the digital city, the environment as surroundings may or may not coincide with the environment as navigation space. We will not, however, distinguish these two environments for the purpose of this study: the latter is often part of the former, and in both cases, the environment is "that, which is not the digital city."

Each user's knowledge is supposed to be a subjective reconstruction of the locally and selectively perceived environment (for justification, see [20, 38]). No user possesses perfect knowledge, but being connected by means of the digital city, the users can gain access to "collective knowledge" – once sensed or created information about the environment that, owing to the spatio-temporal dynamics uniquely allocating each user (and yet the natural cognitive limitations), is far more complete and encompassing than knowledge of a solitary user.

Perception is, obviously, effective only when it provides the navigation process with comprehensible and meaningful – useful – information. This requirement defines the strategy for a digital city. To ascertain the usefulness of a particular perception, the digital city puts it into the context of a situation associated with a user's query and then attempts to evaluate the user's reaction and/or feedback. There can be different and even conflicting interpretations of the same situation made by different users that would, in the long run, destroy the digital city by denouncing its very rationale. In order for a digital city to "survive," its functionality is kept up by enabling context-sensitive (i.e. dependent on the user's prior experience and personal understanding of the situation) interpretation of its contents. The latter sets conditions for the tactics. Thus, the global organizational stability of a digital city (that actually determines its functional stability that is supporting navigation) is naturally maintained at the expense of the stability of its parts (i.e. at the expense of the uniformity of the representation and understanding of the environment – see, for instance, the adaptive interface concept for the Kyoto Digital City described in [17]), just like as it happens in physical cities [30].

It is now understood that a digital city can become self-organizing, if it separates itself from the environment by developing an eventually autonomous structure allowing for generating meanings – forming a "sense" – of the environment for the needs of navigation. In addition to plain separations in matter (there is not such a thing as information, but a digital city is a representation of things; on the other hand, "the environment contains no information; the environment is as it is" [10]) and time (the time-scales of a digital city and its environment usually differ), a self-organizing digital city should develop a meaning boundary: it should maintain and reproduce its own functionally invariant meaning-making structure not just by storing some observations, but by recursively producing pertinent observations using other observations, while acting independently (perhaps, to a degree) of environmental disturbances.

# 3 Approach

## 3.1 Semiotics

An issue of great (by its consequences) significance that is sometimes overlooked by theorists and, fairly often, by practical developers dealing with human-computer interface is the fact that information transmitted by means of computers tends to loose its meaning. Digital signals, such as arrays of bits forming raster graphics, do not bear semantics and have to be interpreted subjectively. There is no context-independent or "absolute" meaning, but the meaning of a signal emerges through the process of interaction between a local perception of the signal and a global (in some way) vision of the corresponding situation [1, 8].

Contrary to the objectivism dominating AI research, human navigation in an environment builds on information conceived (not just perceived!) by the navigator [34]. For instance, observing a map is useless for the purpose of navigation unless the map can be related with the navigator's current location and goal that, as a rule, requires additional "information processing," such as (re)interpretation of sensed information about the surroundings and the map itself. People, through their activities and practice, subjectively and locally but always internally create meanings of the environment. These meanings are then "externalized" to be disseminated and proliferated, while their validity (in respect to the environment) is continuously and again subjectively examined in an attempt to identify currently effectual and supportive meanings. Semiotics studies these in essence meaning-making processes, construing elements of the environment as signs that need to be interpreted to obtain meanings for their contextual use.

In Peirce's formulation [28], semiotics studies the process of interaction of three subjects: the sign itself – the representamen or signifier, the object – that which is signified by the sign; and the interpretant – the meaning that follows semantically from the process of interpretation of the sign. It is postulated that no sign is directly connected to an object: signs have meanings only when they are embodied into a system of interpretance that is just a (larger) system of signs – a sign system, which constrains and relates its constituents, thereby creating a context. A representamen is necessarily a sign of an object for a certain sign system but not for any sign system: depending on the context, the same sign may have different meanings while signifying different objects, or different signs may have the same meaning while signifying the same object, and so on. Designated *semiosis processes* determine the meaning(s) of signs in all the specific situations.

The science of semiotics has a long history of development and application, presently offering a set of generic concepts and procedures to a variety of disciplines, such as art theory, film theory, linguistics, theoretical biology, complex system theory, anthropology, and philosophy of mind, among others (see [6] for a gentle introduction into the study). Often thought of as "the mathematics of humanities" [2], semiotics has developed analytical apparatus for qualitative characterization of various representation and re-representation processes involving signs. This has later been applied on a more formal basis in natural sciences, putting forward a common language for treating information-processing aspects of inter-disciplinary problems [9]. In computer science, semiotics has traditionally been focused on analyzing the reciprocal influence of the computation and interpretation processes and classifying representations by type of relation to their objects [2, 11]. From a semiotic point of view, many (if not all) information processes in a digital city – from "purely" technical, such as data storage, to experiential and cognitive, such as understanding of data – are semiosis processes [36]. Semiotics appears particularly apt for explicating the structure of the mechanism "producing" *meanings out of perceptions*.

### 3.2 Structure for Producing Meaning

Semiotics teaches us that people perceive an environment through signs, which may be interpreted and which serve to mediate meanings of the environment. Although human perception is relatively uniform and consistent, the meaning assigned to a single sign can vary significantly, resting on the subjective dynamics of perception and cognition, as well as on a larger context (e.g. orientational, functional, or operational) of the situation encountered. A semiosis process is the process of determining the meaning of some distinctions in an environment that entails representation and re-representation of these distinctions over several levels of interpretation, each of which is governed by and adopts certain *norms* – developmental rules and relational constraints for the signs. The norms reflect different aspects of human behavior that can be classified into five major groups [35]: perceptual – to respond to peculiarities of sensing; cognitive – to deal with cultural knowledge and beliefs; evaluative – to express personal preferences, values, and goals; behavioral – to delineate behavioral patterns; and denotative – to specify the choice of signs for (further) signifying.

From a system-theoretic viewpoint, the complexity and richness of many natural organizational processes, such as adaptation and self-organization, derives from the ability to arrange smaller units into larger ones, which are in turn arranged into larger ones, which are arranged into still larger ones, and so forth [33]. Semiosis is a natural organizational process [21]: it organizes signs in a partial hierarchy by ordering them so that representamina of objects (that can be other signs) of level N-1 for processes and structures of level N+1 are placed on level N. The lowest-level signs, e.g. (manifestations of) physical objects, behavioral dispositions, emotions, and the like, are perceived or realized through their distinctions and get a representation at an "intermediary" level of norms, reflecting interpretive laws of a higher, experiential and environmentally (physiologically, socially, technically, economically, etc.) induced level, which accommodates interpretants and gives meanings to the representamina. This simple three-level structure corresponds to and is set up by a single semiosis process, whereas various semiosis processes defined on the same realm will create a complex partially ordered structure, where one sign gets multiple meanings, depending on both the signified contents of the lower levels and the contextual constraints from the interpretive levels (see Fig. 1).

Navigation with a digital city activates a number of semiosis processes (e.g. by different users) and results in the creation of a multi-level sign system with a potentially infinite hierarchy of interpretive levels, where signs on level N are dynamically com-posed of signs on level N-1 so that only those of all the possible combinations of that lower-level signs persist, which are allowed by boundary conditions effective at level N+1. Signs on level N-1 serve as constitutive units for level N signs, which are constitutive for level N+1 signs, which can be constitutive for yet-higher-level signs; besides, signs on higher levels are constraining for signs on their adjacent lower levels. The levels have different dynamics, such that the probability of changing the relationships among signs within a level decreases for higher levels [21].

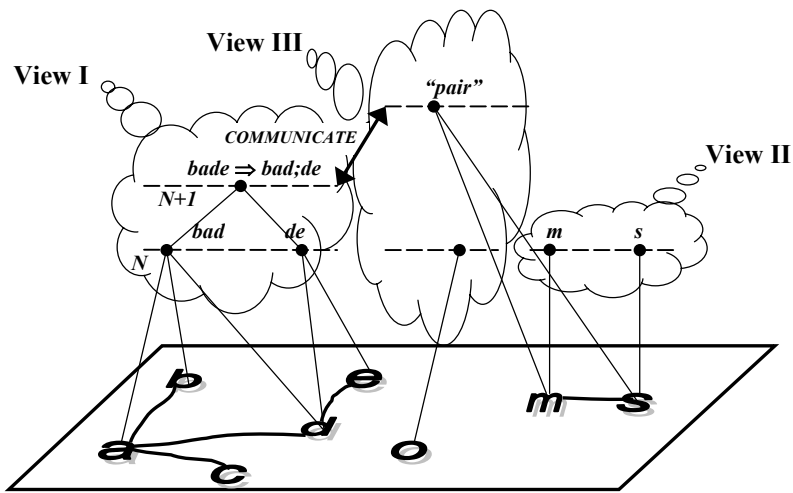

**Fig. 1.** A simplistic example of the composite partial hierarchy for producing meaning: in View I, entities "***b***, ***a***, ***d***" and "***d***, ***e***" are represented at level *N* as *connected* for level *N+1*, and a new (in respect to *N*) meaning may emerge (e.g. be inferred) for *N+1*: "***b*** and ***e*** are connected." There are some other systems of interpretance in the structure, and the one that recognizes, for instance, *pairs* of connected entities – as in View III. By referring to (i.e. communicating with) III, View I may apparently learn (e.g. *syntactically* – see [38]) to also recognize pairs of entities. It may then reconsider "***bade***" as a pair "***bad;de***" (that will become of interest to III). Thus, the meanings in the system may change as I and III communicate.

A user of a digital city typically deals with a fragment of the global, i.e. loosely shared through the environment (that may be seen as the lowest semiotic level) by all the users, system of signs (Fig. 1). The fragment is, however, distinctively ordered in an interpretive hierarchy peculiar to the user's experience and the norms he or she adopts. Hierarchies created by different users may be different in terms of the order as well as the coverage, and they may run on different time-scales (see [20] for general argumentation). Having been combined into one structure (e.g. by means of a community network [12]), the fragments may form a global but partial and often implicit "hierarchy." This global hierarchy constitutes the functionally invariant structure of a self-organizing digital city. It allows for producing "meanings" for the system internal (adaptation) needs out of (represented) perceptions based on experience (received through, for instance, feedback – see Section 2.2) currently prevailing in the society of digital city users. The hierarchy has essentially an ordering, i.e. affecting the interpretive levels rather than signs within a level, dynamics [27].

Unlike the case of individual navigation, where perceived and conceived signs may need not be articulated explicitly, the development and operation of a digital city neatly builds on communicative use of a multi-level sign system representing the environment and the digital city itself. This sign system can be externalized – derived from the digital city structure – as a language defined in a very general way and not confined to handling verbal constructions. The digital city "describes" (and interacts with) its environment with this language, which has a syntax reflecting the organization of the environment, semantics defining meanings of the environment, and pragmatics characterizing the effect of the language use. (See, for instance, [12] presenting the Campiello System that works to construct and utilize such a language.)

## 4  Self-Organizing Digital City

### 4.1  Communication

Users of a digital city, although act individually, are not isolated from the surroundings: their behavior is determined not only by their purposes, but considerably by the material processes taking place in the environment, the actions of other individuals (both, users and non-users), the existing time- and functionality-constraints, the actions of groups of other individuals, the current state of the digital city, and the like. The users interact with the digital city (yet being parts of it), their environments, and simply with each other. The users are, nonetheless, autonomous, in the sense that they possess a representation of the environment adequate to sustain their purposeful behavior for some time, as autonomous is the digital city, which is recursively (through its users) closed with respect to meaning. In this situation, the operation of a digital city heavily depends on the social context of the semiosis associated with the navigation process – it depends on how the users and the digital city receive and interpret information in the course of navigation, i.e. how they communicate.

From a behavioristic point of view, an individual engaged in navigation develops an internal representation using those distinctions of the environment, which turn up solutions to the problem that are successful behaviors [5]. Signs of such a representation arrive as "tools for indication purposes" [32]. When met in an environment, these signs (i.e. the distinctions they stand for) serve to orient the navigator, whatever their "actual" meanings or rôles could be in the environment. The navigator is not really interested in "getting to the truth," but in knowing what happens or what are possible consequences – *expectations*, when a sign is encountered. In this aspect, signs are signifiers emerged of successful interaction between an individual and an environment as orientational "pointers" to not just an object standing in a referential relation with the sign, but to the outcomes desired for (or, at least, anticipated by) the user. Signs can be considered "anticipations of successful interactions of referral" [32], emphasizing their origin and predictable influence on behavior.

One can show that the behavioristic view of the foundational process of forming sign meanings is just a specialization of the classical view that defines information as "a difference that makes a difference" to the interpreter [4]. The specialized view, however, makes it difficult to explain communication as mere exchange of signs. Indeed, in the case of navigation with a digital city, not objective reality but subjec-tive experience is the grounding basis for signs (also see discussion in [19]). The navigator cannot frequently succeed with developing an interpretation of a sign received through communication by simply referring the sign to the observed part of the environment – the navigator's personal experience has first to be "synchronized" (up to a point) with the experience underlying the creation of the sign. The latter appears impossible or inefficient (e.g. because of time-limitations) in most cases of the use of a digital city.

A solution to the above problem comes with an advanced explanation of communication that includes aspects of information (sign) exchange as well as behavioral coordination between *autopoietic systems*. An autopoietic system is a dynamic system maintaining its organization on account of its own operation: each state of such a system depends on its current structure and a previous state only [24]. The structure of an autopoietic system determines the system possible (i.e. self non-destructive) behaviors that are triggered by its interactions with the environment. If the system changes its state, enforcing changes of the structure without breaking autopoiesis, the system is *structurally coupled* with the environment. If the environment is structurally dynamical, then both the system and the environment may mutually trigger their structural changes, sustaining the system's *self-adaptation*. When there are more than one autopoietic system in the environment, the adaptation processes of some of the present systems may become coupled, acting recursively through their own states. All the possible changes of states of such systems, which do not destroy their coupling, create a *consensual domain* for the systems. Behaviors in a consensual domain are mutually oriented. Communication, in this view, is the (observed) behavioral coordination resulting from the interactions that occur in a consensual domain (see [8] for details).

(Human) users of a digital city are (higher order) autopoietic systems [24]. Besides, the environment of a digital city is supposed to be structurally dynamical or even self-organizing (to a degree), as in the case of social systems [14, 23, 38]. Therefore, a digital city should be autopoietic (at least, to a degree), i.e. to be a system *internally* producing meanings for its own (adaptation) needs, to endure communication.

### 4.2 Definition of digital city

Based on the system-theoretic and semiotic analysis made in the previous sections, we can now define a digital city as follows:

*A digital city is an autopoietic organization of social agents communicating by way of computers, such that every social agent is a realization of a semiosis process engendered by navigation taken place in a common (for all agents) environment.*

It is important to notice that the above definition builds on the understanding of communication as the (observed) coordination activity in a consensual domain, and it does not "humanize" the social agency: equally, people and computer (and any other) systems can be social agents as long as they interact and produce meanings in the navigation process; besides, the term "navigation" is understood in the broad sense, following [34].

It can be seen that the proposed definition is general enough to encompass all the realizations of digital cities reported in the literature, which have a common identity, constituting a distinct class of digital products. On the other hand, it is sufficiently precise in giving not only the functional (i.e. *what* goes on) but operational (i.e. *how* it goes on) characterization of a digital city. By the definition, an agent does not have to physically be embedded into the navigation space but does have to be engaged into the navigation process. The latter allows us to clearly distinguish a digital city among other web-based digital products, yet leaving plenty of freedom for dynamically including and excluding resources and agents appearing in it. One should not, nevertheless, be confused by the process-orientation of the definition: not every navigation (e.g. on the World Wide Web – see [14]) is the source of the emergence of a (self-organizing) digital city, but only that, which is supported by (and supports) the functionally invariant structure for navigation in the specified space.

## 5   Discussion and Conclusions

Within descriptions of digital cities, there is often little attention to the precise definition of basic concepts, with which a digital city is characterized. This imprecision results in weakly motivated developments, which easily loose their identities when compared with other digital products, such as map repositories or Web-portals (consider, for instance, the Turin and AOL digital cities discussed in [18]). Moreover, although it is assumed by default that a digital city is deployed for a group of users rather than for a single user, most of the reported projects habitually focus on and address specific aspects related to the personal adaptation (e.g. of the interface), while the issue of the appropriateness of a digital city to a particular society remains opaque. Even less is known about possible mutual influences of a digital city and the society, and about the life cycle of a digital city. All this could be a serious reason to question the very expediency of digital cities.

In this paper, an attempt was made to find a theoretical basis for the development of digital cities. Starting from an assertion that "digital city" is a metaphor called to denote a complex digital product with properties structurally similar to the ones of physical cities, the concept of digital city was gradually refined throughout the study, as we analyzed it first – functionally, then – semiotically, and finally – from a system theoretic perspective. The definitive function of a digital city is the (information) support of the process of navigation in an environment. Navigation utilizes meanings of the environment resulting from perception and interpretation. Interpretation is intrinsic of semiosis. Different meanings are developed by semiosis processes, which create and order signs of the environment into partial hierarchies. Semiotic sign-hierarchies emerged during navigation can internally generate new semiosis processes and, therefore, new meanings owing to communication. If this generation is maintained regardless environmental variations, the organization of semiosis processes becomes autopoietic, and it constitutes a digital city.

The authors are quite aware of the difficulties and controversy attributed to any attempt to define a "not obviously living" system as autopoietic. By arguing that a digital city should be autopoietic, we follow the German sociologist and philosopher Niklas Luhmann [23], who was first to explicate the autopoiesis of social systems. The concept of digital city is, in our opinion, to organically expand the communication-driven autopoiesis of social systems to the new "digital" dimension. It should be stressed that neither semiotic meaning-making nor self-organization alone is an entirely new and unexplored issue in the fields of Human-Computer Interaction and Computer Supported Cooperative Work (see, for instance, [1] and, especially, [26]). Somewhat different to the previous works, we see the advantage of the application of semiotics and complex system theory not only in their suitability for theoretical exploration of digital products, but in their appropriateness for a rigorous computational treatment and technological validation of the theoretical findings, as it became apparent with the recent advent of algebraic semiotics and category theory [11], as well as evolutionary computation [8].

The view of digital city developed through the study is not only fully compatible with the contemporary vision of urban communities as self-organizing systems (a city as "a text written by millions of unknown writers…" [29, 30]), but it specializes and details the mechanism of self-organization by advocating that digital cities are autopoietic. By extending the recently popular idea that not just biological, but also psychic and social systems can be autopoietic [23, 24, 14], autopoiesis can be considered as a general form of system development that draws on self-referential closure [9, 15]. It was argued in the paper that in the case of digital cities, the concept of life, which is exploited in biology and (in a sense) sociology and urbanism, is to be replaced with the concept of semiosis as a kind of autopoietic organization. Along with systems theory that is used in the study of urbanism [30], semiotics forms the basis for investigation of (self-organizing) digital cities.

Semiosis of a digital city arranges a structure required for the reproduction of meanings by the digital city for its own "internal" needs. The meaning-(re)producing autopoiesis gives more room for a system to "survive" by letting it be autopoietic to a degree: a digital city can be less (and trivially) autopoietic if it mainly produces meanings out of perceptions, and it can be more autopoietic if it produces meanings out of meanings. For the former, consciousness of the users is the source of meaning reproduction that is typically fairly dependent on the environment, i.e. on what is not the digital city. For the latter, meanings are reproduced in the course of communication that powers the autopoiesis without paying much attention to the environment. "Meanings out of perceptions" assumes a hetero-referential closure of the digital city: the system produces meanings for other (possibly "external") meanings (e.g. it learns to send an image for clarifying a text). "Meanings out of meanings" implies a self-referential closure: the digital city produces meanings for communication (e.g. it learns to send an image for adjusting its own interface). Resembling social systems [23], self-reference for a digital city is the ability to distinguish between hetero-reference and self-reference.

The proposed definition of digital city departs from the criticized studies (also see [38] for a more general critique), which tend to focus on the micro-scale phenomena concerning the interaction of a user with a (part of a) digital city but ignore (or artifi-cially "fix") the global social dynamics of the digital city. All the users, through their social agents that are realizations of semiosis processes, are constitutive parts of the digital city, and its dynamics is determined by behaviors of the users. It is obvious that the concept of digital city becomes incongruous if the users are not included into the definition in the case of hetero-referential closure: the system would then be anything – from a database to a game –, depending on the purpose of the user, i.e. on what is the motivation or "driving force" causing the system of interpretance for the semiosis processes representing the user in the interaction with the digital city. Besides, the concept is just absurd if devoid of the users in the case of self-referential closure: how to call a digital product, which acts for its own purposes, leaving an external user unaware of them, and which is generally unpredictable in its behavior? Contrasting these two extreme points, the global and inclusive treatment advocated in this paper is not only comprehensive, but it allows for applying the rich apparatus of social studies to the study of digital cities to examine the macro-phenomena. At the same time, the proposed approach well recognizes the micro-scale dynamics: any user, whether an individual or a group, can uniquely be defined through characteristic semiosis processes. This gives us a happy opportunity to apply various theories of human-computer interaction as well as semiotics to the research and development. Combining the micro- and macro-level visions, the meaning-making self-organization implies the emergence of some ontology (following the terminology coined by the knowledge-sharing research community [13]) of the navigation space, which is understood as the functionally invariant structure of a digital city. This ontology, however, has a structural dynamics and changes throughout the life cycle of a digital city. New technological perspectives might be discovered when examining this evolution (e.g. how networking and information retrieval mechanisms should react when the digital city undergoes a change from hetero- to self-referential closure), allowing for "deliberately participating" in the digital city development (compare with the study of urbanism, [30]) that would lead us to the creation of a "participative virtual city," e.g. as discussed in [3].

The presented work offers one new contribution: the clarification of the concept of digital city. This contribution is based upon the extensive analysis supported by the literature. Another contribution of the paper would thus be providing the reader with an introduction (though, by no means complete) to the semiotic and system-theoretic aspects of the study of social system.

We do not expect our approach be perfect. The presented study seeks to explain a particular view of digital cities that would be found somehow inappropriate. We believe, however, that this is better than discussing the subject in such an elusive way, that no one can tell if it is inappropriate. This work also is to stimulate critical discussions of the concept of digital city.

Building on the conceptual and terminological basis developed through the study, our future research plans include: 1) elaboration of a semiotic theory of communication for a digital city, 2) its verification by both analysis of practical examples and computational experiments, 3) exploration of possible implications of the study of digital cities for the study of urbanism.

## Acknowledgement

The presented work is part of the Universal Design of Digital City project in the Core Research for Evolutional Science and Technology (CREST) programme funded by the Japan Science and Technology Corporation (JST).